\documentclass[10pt,journal,compsoc]{IEEEtran}

\usepackage{graphicx}
\usepackage{amsmath,amssymb} 
\usepackage{color}
\usepackage{amssymb}
\usepackage{mdwmath}
\usepackage{mdwtab}
\usepackage{amsmath}
\usepackage{epsfig}
\usepackage{graphicx}
\usepackage[font=small]{caption}
\usepackage[font=small]{subcaption}
\usepackage{tabularx}
\usepackage{multirow}
\usepackage{algpseudocode,algorithm,algorithmicx}
\usepackage{float}
\usepackage[pagebackref=true,breaklinks=true,,colorlinks,bookmarks=false]{hyperref}

\ifCLASSOPTIONcompsoc
  \usepackage[nocompress]{cite}
\else
  \usepackage{cite}
\fi
\ifCLASSINFOpdf
\else
\fi
\begin{document}

%
\title{Deep Learning of Appearance Models for Online Object Tracking}
%
%
%
%

\author{Mengyao Zhai,
        Mehrsan Javan Roshtkhari,
        Greg Mori
        }
\IEEEtitleabstractindextext{%
\begin{abstract}
This paper introduces a novel deep learning based approach for vision based single target tracking. We address this problem by proposing a network architecture which takes the input video frames and directly computes the tracking score for any candidate target location by estimating the probability distributions of the positive and negative examples. This is achieved by combining a deep convolutional neural network with a Bayesian loss layer in a unified framework. In order to deal with the limited number of positive training examples, the network is pre-trained offline for a generic image feature representation and then is fine-tuned in multiple steps. An online fine-tuning step is carried out at every frame to learn the appearance of the target. We adopt a two-stage iterative algorithm to adaptively update the network parameters and maintain a probability density for target/non-target regions. The tracker has been tested on the standard tracking benchmark and the results indicate that the proposed solution achieves state-of-the-art tracking results.
\end{abstract}

}

\maketitle

\IEEEdisplaynontitleabstractindextext

%
\IEEEpeerreviewmaketitle

\IEEEraisesectionheading{\section{Introduction}\label{sec:introduction}}

Visual target tracking is a fundamental task in computer vision and vision based analysis. High level video analysis applications typically require the objects of interest to be tracked over time. Single object tracking is a well researched topic for which a diverse set of approaches and a rich collection of algorithms have been produced to date. Tracking is an almost solved problem when objects in a scene are isolated and easily distinguishable from the background.  However, the problem is still challenging in real world applications because of occlusions, cluttered backgrounds, fast and abrupt motions, dramatic illumination changes, and large variations in the viewpoint and pose of the target.  Readers may refer to ~\cite{YangSFL2011} and ~\cite{WuLimYang13} for a review of the state-of-the-art in object tracking and a detailed analysis and comparison of representative methods. 

In general, single target tracking algorithms consider a bounding box around the object in the first frame and automatically track the trajectory of the object over the subsequent frames. Therefore, single target tracking approaches are usually referred to as ``generic object tracking'' or ``model-free tracking'', in which there is no pre-trained object detector involved~\cite{Collins2005online,ZhangCVPR15Structure,Ross2008incremental,Henriques2015KCF}. Model free visual object tracking is a challenging problem from the learning perspective, because only a single instance of the target is available in the first frame and the tracker must learn the target appearance in the subsequent frames. In almost all of the previously reported algorithms, the object itself and/or its background are modeled using a local set of hand-crafted features. Those features can be based either on intensity or texture information~\cite{HareSTRUCK2011,Zhang2012Compressive,dinh2011context,Kalal2010PNclassifiersTracking} or color information~\cite{Danelljan2014adaptive,Possegger2015color,Oron2012OrderlessTracking,Nummiaro2003ColorParticleFilter}. Those feature vectors are then employed either in a generative~\cite{AdamRSCVPR2006,Kwon2010Decomposition,Jepson2003Robust,Zhang2012Compressive} or discriminative~\cite{Babenko2013-MIL,Avidan2007Ensemble,Collins2005online,Grabner2008Boosting,Kalal2010PNclassifiersTracking,Avidan2004SVMtracking} inference mechanism in order to detect and localize the target in the following frame. It has been demonstrated that the most important part of a tracking system is representative features~\cite{ZhangZY2013}. Although the currently used hand-crafted features produce acceptable tracking results, it is always preferred to leverage more descriptive features. Therefore, it is more beneficial to exploit target-specific representations through a learning process rather than using a fixed set of pre-defined features~\cite{KuenPATREC2015}.  

Deep neural networks and in particular Convolutional Neural Networks (CNN) have recently gained a lot of interest in computer vision due to their strong capabilities in various applications such as object detection~\cite{girshick14CVPR} and classification~\cite{NIPS2012_4824}. However, these techniques are not well exploited in visual tracking because of the inherent difficulties for deep neural networks in learning from small numbers of positive examples. In addition, it seems that deep learning architectures are better in learning category specific features rather than target specific ones; further, they usually require a long iterative training process to converge, which makes it difficult for them to be used in an online learning procedure. Therefore, extending current deep learning architectures to the tracking problem is not a straightforward process. Recent work has pursued attempts to use them for tracking, showing some promising results~\cite{NIPS2013_5192,li122014deeptrack,li2015robust,wang2015transferring,wang2015video,hong2015online}. In this paper we propose an online object tracking algorithm based on deep neural networks, in which the network learns the probability densities of appearance of the target and its surroundings and updates itself adaptively to new observations.

\begin{figure*}[!ht]
\begin{center}
\includegraphics[width=0.95\linewidth]{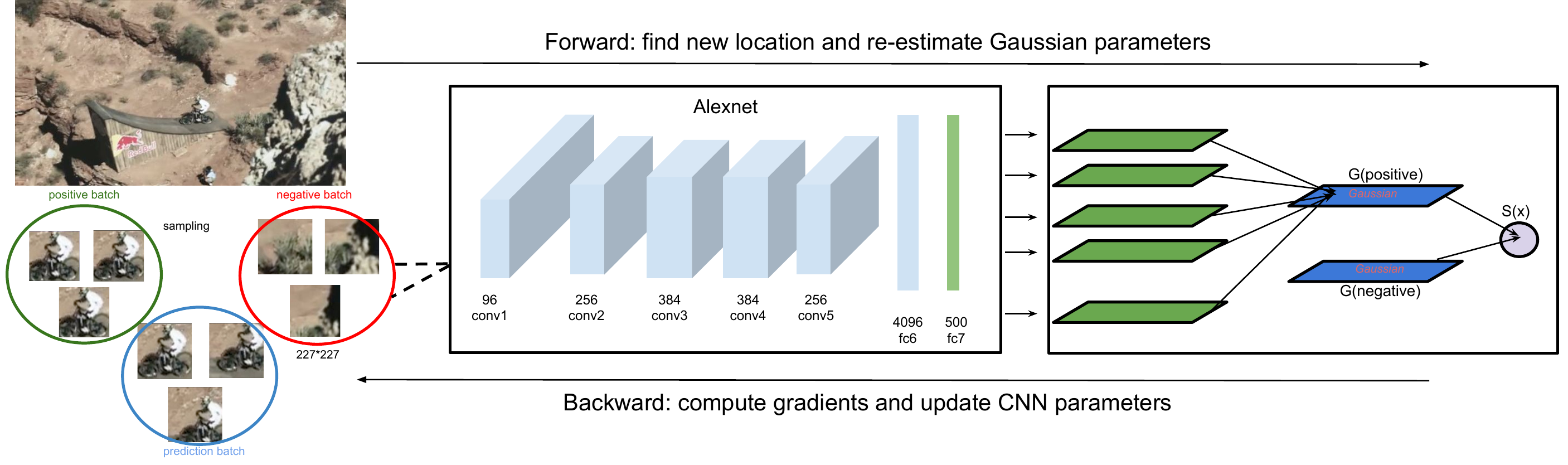}
\end{center}
 \caption{Overview of our approach: Given one frame, we sample three batches: positive batch, negative batch, and prediction batch. In the forward procedure, given CNN parameters, we use the positive batch and negative batch to re-estimate Gaussian parameters. Then we search in the prediction batch for the new location with maximum score. In the backward procedure, given Gaussian parameters, we compute gradients with respect to feature nodes and update CNN parameters.}
\label{fig:TrackingOverview}
\end{figure*}

\figurename{~\ref{fig:TrackingOverview}} shows the overall architecture of the proposed tracking system. The system consists of a two stage training process, an offline training procedure and an online target specific fine-tuning step. The first step is carried out by taking a pre-trained CNN which is already trained for large-scale image classification tasks, and then is fine-tuned for the generic object detection task which is referred to as \emph{objectness}~\cite{AlexeDF10}. The network is then fine-tuned based on the target appearance in the first frame.
This results in a network that can discriminate the target from the background. 

The target detection and classification is performed by adding Gaussian distributions to the fc7 layer in order to estimate distributions of the positive (target related) and negative (background related) representative features learned by the network. The final decision is made by a naive Bayes classifier to estimate the true location of the target in the following frame. The online training process uses a two-stage iterative algorithm to adapt itself to target appearance changes by updating both network and Gaussian parameters. During the online training process, the Gaussian parameters are updated during a forward pass given a set of fixed network parameters, and then the target detection error is back-propagated to update the network parameters given a set of fixed Gaussian parameters. The Gaussian and network parameters are updated accordingly in each frame.  

In this paper, we propose a new deep learning based tracking architecture that can effectively track a target given a single observation. The main contribution of this paper is a unified deep network architecture for object tracking in which the probability distributions of the observations are learnt and the target is identified using a set of weak classifiers (Bayesian classifiers) which are considered as one of the hidden layers. In addition, we fine-tune the CNN tracking system to adaptively learn the appearance of the target in successive frames. Experimental results indicate the effectiveness of the proposed tracking system.

\section{Related Work}
\label{sec:previous_work}

To date, most of the reported tracking approaches rely either on robust motion or appearance models of each individual object using hand crafted features. Recently, some approaches have been introduced to take advantage of the CNN's feature learning capabilities for visual tracking. In general, there are two different ways to employ deep neural networks for tracking; one is to use them as a feature extractor combined with a discriminative classifier~\cite{NIPS2013_5192,FanXWG2010} and the other is to use a whole neural network tracking pipeline~\cite{NIPS2013_5192,wang2015transferring}.

One of the early methods for deep learning based tracking has been introduced in~\cite{NIPS2013_5192}, in which a stacked denoising autoencoder network is employed to extract image features and then a particle filtering approach is used to track the target. The network is pre-trained using auxiliary data (80 million samples from the Tiny Image dataset~\cite{Torralba2008TinyDataset}) in order to learn generic image features, which are transferred for online tracking. While this approach learns features for tracking, it is preferable to learn target-specific features in a tracking task rather than use generic image features. The simple architecture of the network and the absence of target-specific fine-tuning hinder the capabilities of this tracker significantly. 

Other approaches are introduced in ~\cite{li122014deeptrack} and ~\cite{li2015robust} by leveraging multiple shallow convolutional neural networks (two convolutional and one fully connected layers) to learn representative features from multiple image cues, i.e. three locally normalized gray scale images and one gradient image. One of the drawbacks associated with these approaches is the limited learning capabilities due to the shallow network structures. The authors have proposed the use of a \emph{pool} of convolutional neural networks with a temporal adaptation mechanism in order to learn all the appearance observations of the target in the past; which limits the application of this approach for long video sequences. 

Another interesting deep learning based tracker is presented in~\cite{hong2015online}. This approach takes advantage of a pre-trained CNN for large scale image classification combined with a Support Vector Machine (SVM) as a target/background classifier in order to generate a saliency map for the target location. The CNN is treated as a permanent image feature representation which remains unchanged during tracking. The tracker adaptation for the target appearance changes is solely handled by the discriminative properties of an online SVM. Here the difficulties for updating the CNN parameters with a limited number of observations are simply avoided by completely ignoring the need for the online fine-tuning, similar to~\cite{NIPS2013_5192}.

Two recently published deep learning based approaches~\cite{ma2015hierarchical} and~\cite{Qi2016HedgedDT} both use CNN features from multiple layers instead of focusing on features extracted from last layer and correlation filters on each convolutional layer are learnt to encode target appearance. In~\cite{ma2015hierarchical}, maximum response of each convolutional layer is inferred to locate the targets. While in~\cite{Qi2016HedgedDT}, a strong tracker is integrated from weak trackers from each convolutional layer by using an online decision-theoretical Hedge algorithm. Another recently published tracker~\cite{wang2015visual} also take advantage of CNN features extracted from different layers. Instead of integrate features of all layers, it used a selection method to filter out noisy, irrelevant or redundant feature maps from two pre-defined convolutional layers.

More closely related to our approach is the one presented in~\cite{wang2015transferring}, in which the authors have developed a structured output CNN for the single target tracking problem. The algorithm processes an input image and produces the probability map (aggregated over multiple scales) of every pixel that belongs to the target that is being tracked. It is then followed by an inference mechanism to detect and localize the target. In other words, their proposed architecture for deep network based tracking produces a pixel-wise probability map for the targetness and hence, the network is considered as a generative model which estimates the likelihood of the target. Their approach consists of two CNNs with different fine-tuning strategies during online tracking which are collaborating together in order to learn the likelihood map for the target presence. Alternatively, in this paper we propose a classification-based tracking scheme in which rather than assigning a targetness probability to each pixel, the target is being identified as a region which has the maximum classification score given the learned models for both positive and negative regions. We employ a Bayesian classifier as a loss layer in our CNN tracker and update the network parameters in online tracking in order to account for the target appearance variations over time.

\section{Tracking With Convolutional Neural Networks}
\label{sec:TrackingMethod}
This section presents the algorithmic description and the network architecture for the proposed tracking system. Considering the network architecture presented in \figurename{~\ref{fig:TrackingOverview}}, the objective is to train a discriminative model to learn the target appearance given a single observation and distinguish it from the background. Given an input image, the output of the tracking system is a classification score of a discriminative appearance based classifier. $\mathcal{S}(\mathbf{x})$ defines the classification score of an observation, $\mathbf{x}$, to be considered as a target. In this section we first describe the network architecture and the target classifier, $\mathcal{S}(\mathbf{x})$, which serves as a loss layer in the network architecture. Then we describe how a pre-trained network is fine-tuned to adapt itself to the target appearance variations and to learn the target appearance in a discriminative manner sequentially by an online fine-tuning strategy. 

\subsection{CNN-Tracker: Architecture Overview}
\label{sec:NetworkArchitecture}

The network architecture is presented in \figurename{~\ref{fig:TrackingOverview}}. This network adopts a CNN model similar to \emph{AlexNet} as its basis, described in~\cite{NIPS2012_4824}. The only difference is that the dimension of fc7 layer is reduced to 500 dimensions. This network consists of five convolutional layers and two fully connected layers. The inputs of the CNN are image patches with the size of $256 \times 256$ pixels. We adopt the pre-trained model, which is trained for image classification and hence, needs to be fine-tuned for tracking. The fine-tuning procedure is explained in more detail in Section~\ref{sec:offlineTuning}.

We denote with $\mathbf{x} = [x_{1}, x_{2}, ..., x_{N}]^T$ the vector representing the output of the second fully connected layer (fc7) in the CNN (highlighted in green in \figurename{~\ref{fig:TrackingOverview}}). It can be considered as a feature vector representing each image patch. Given the probability distributions of the negative and positive examples, the discriminative classifier for target detection and localization can be modeled using a naive Bayes classifier:
\begin{equation}
  \mathcal{S}(\mathbf{x}) = log\left( \frac{P(\mathbf{x}|pos) P(pos)}{P(\mathbf{x}|neg)P(neg)}\right)
  \label{eq:BayesClassier}
\end{equation} 

We assume that the prior probabilities of the positive and negative labels are equal and features are independent. Then~\eqref{eq:BayesClassier} is rewritten as\footnote{The same assumption is made in other tracking methods, such as the MIL-Tracking algorithm~\cite{Babenko2013-MIL}.}:
\begin{equation}
  \mathcal{S}(\mathbf{x}) = log\left(\prod_{i=1}^{n}{\frac{P(x_i|pos)}{P(x_i|neg)}}\right)
  \label{eq:TrackerClassier}
\end{equation}

We assume that the distributions of the positive and negative examples' features can be represented by Gaussian distributions. Similar to~\cite{Babenko2013-MIL,ZhangZY2013}, it is assumed that the distribution of the posterior probability of the positive examples, $P(\mathbf{x}|pos)$, obeys a single Gaussian distribution denoted by $\mathcal{G}_{pos}$ and the posterior distribution of the negative examples, $P(\mathbf{x}|neg)$, 
obeys a Gaussian distribution\footnote{A single Gaussian model can be replaced by Gaussian mixture models to capture the diverse appearances and shapes of the negative examples more effectively.}, $\mathcal{G}_{neg}$. We use single Gaussian model for simplicity and experimental results in Section~\ref{sec:results_part2} suggest that single Gaussian is powerful enough in this case. Therefore:
\begin{align}
  \mathcal{G}_{pos} &=P(\mathbf{x}|pos)
   \notag \\
   &=\prod_{i=1}^{N}\frac{1}{ \sqrt{2\pi}\sigma_{{pos}_i}}e^{-\frac{(x_{i} - \mu_{{pos}_i})^2}{2{\sigma^2_{{pos}_i}}}}
  \label{eq:positivesGMM}
\end{align}
where ${\mu_{{pos}_i}}$ and ${\sigma_{{pos}_i}}$ are the mean and variance of the Gaussian distribution of the $i^{th}$ attribute of the positive feature vector, $x_i$, respectively. Similarly, we can get distribution for negative examples.

\subsection{Initialization: Two Phases of Network Fine-Tuning}
\label{sec:offlineTuning}

The fine-tuning of the pre-trained network is carried out through two phases: \emph{obj-general} as phase 1 and \emph{obj-specific} as phase 2. The pre-trained network is originally trained for image classification and hence, it does not suitably represent the appearance of a specific target. Therefore, it cannot be used directly for a tracking task. For object tracking tasks, the objective is to precisely localize the target in a given frame. In practice, there are not enough examples available to train a deep network to recognize and localize the target. To address this problem, given a pre-trained model, we fine-tune the network twice in order to reach our final goal: learning the appearance of the specific target of interest, given a single example. At phase 1, we try to learn generic features for all object types and given CNN unchanged, we start phase 2 fine-tuning by only replacing the data we use so that the network can focus on learning features for specific object.

 In order to learn generic features for objects and be able to distinguish objects from the background, we sampled $100k$ auxiliary image patches from the ImageNet 2014 detection dataset\footnote{\url{http://image-net.org/challenges/LSVRC/2014/}}. For each annotated bounding box, we randomly generate negative examples from the images in such a way that they have low intersection of union with the annotated bounding box. During this phase, all CNN layers are fine-tuned. The fine-tuned CNN can now be considered as a generic feature descriptor of objects, but it still cannot be used for tracking because it cannot discriminate a specific target from other objects in the scene. In other words, this network is equally activated for any object in the scene. 
 
 Another phase of fine-tuning is conducted given the bounding box around the target in the first frame. In order to generate a sufficient number of samples to fine-tune the network, we randomly sample bounding boxes around the original one. Those bounding boxes have to have a very high overlap ratio with the original bounding box. For the negative bounding boxes we sampled bounding boxes whose centers are far from the original one. During this phase, only fully connected layers are fine-tuned.

\subsection{Simultaneous Tracking and Online Network Fine-Tuning}
\label{sec:onlineTuning}
The input to our tracking algorithm is a single bounding box representing the target of interest in the first frame of a video sequence, which can be generated by either running an object detector or using manual labeling. Given that bounding box, first we use a sampling scheme to sample some positive patches around it and some negative patches whose centers are far from positive ones. Therefore, the probability density functions of the positive and negative examples are computed using~\eqref{eq:positivesGMM}. When a new frame comes, we sample some possible locations around the previous location of the target with a predefined search radius\footnote{The search region is a function of similarity of the computed features at the current frames to the previous observations.}.

The sampling is done at multiple scales by building an image pyramid and a set of candidate bounding boxes is generated which is referred to as $\mathbf{X}$. Given a candidate bounding box in the frame, $\mathbf{x}_i \in \mathbf{X}$, the tracking score is computed as: 
\begin{equation}
    \mathcal{S}(\mathbf{x}_i) = \log(\mathcal{G}_{pos}(\mathbf{x}_i)) - \log(\mathcal{G}_{neg}(\mathbf{x}_i))
    \label{eq:TrackingScore}
\end{equation}

The candidate bounding box which has the highest tracking score is then taken to be the new \emph{true} location of the target:
\begin{equation}
    \mathbf{x}^* = \arg \max_{\mathbf{x}_i \in \mathbf{X}} \mathcal{S}(\mathbf{x}_i)
\end{equation}

Once the \emph{true} target bounding box is determined in the following frame, the whole model shall be fine-tuned again in order to adapt itself to the new target appearance. We consider a two-stage process to update the model.

\noindent \textbf{Stage 1: Updating Gaussian parameters}. The means and variances are updated using the new prediction $\mathbf{x}^*$. Given location of $\mathbf{x}^*$, positives and negatives are sampled again. Assume that means and variances for the new positives are estimated as $\mu_{{pos}^*}$ and $\sigma_{{pos}^*}$. Then we update means and variances according to the following equations:
\begin{equation}
     \mu_{{pos}_i} = \gamma \mu_{{pos}_i} + (1-\gamma) \mu_{{pos}^*_i}
     \label{eq:PostivieExamplesUpdate1}
\end{equation}
\begin{align}
    \sigma^2_{{pos}_i} = & \gamma \sigma^2_{{pos}_i} + (1-\gamma)\sigma^2_{{pos}^*_i} + 
    \notag \\
    & \gamma(1-\gamma)(\sigma_{{pos}_i} - \sigma_{{pos}^*_i})^2 
    \label{eq:PostivieExamplesUpdate2}
\end{align}
where $\gamma$ is the learning rate. Similarly, the mean and variance of the Gaussians reprsenting negative examples' distributions can be updated using the same update rules in \eqref{eq:PostivieExamplesUpdate1} and \eqref{eq:PostivieExamplesUpdate2}.

\noindent \textbf{Stage 2: Updating the network weights}. Given feature $\mathbf{x}$ extracted from an image patch, the corresponding tracking score is computed by \eqref{eq:TrackingScore}. Therefore, it is expected that the tracking score is maximized for positive examples while being minimized for the negative ones. In order to update the network weights, the gradients of the $i^{th}$ element of positive $\mathbf{x}$ are computed as follows:
\begin{equation}
    \frac{\partial \mathcal{S}}{\partial x_i} = \frac{\partial (\log(\mathcal{G}_{pos}(\mathbf{x})) - \log(\mathcal{G}_{neg}(\mathbf{x})))}{\partial x_i}
\end{equation}

And the gradients of the $i^{th}$ element of negative $\mathbf{x}$ are computed as follows:
\begin{equation}
    \frac{\partial \mathcal{S}}{\partial x_i} = -\frac{\partial (\log(\mathcal{G}_{pos}(\mathbf{x})) - \log(\mathcal{G}_{neg}(\mathbf{x})))}{\partial x_i}
\end{equation}

where 
\begin{align}
    & \frac{\partial \log(\mathcal{G}_{pos}(\mathbf{x}))}{\partial x_i} = -\frac{x_{i} - \mu_{{pos}_{i}}}{\sigma^2_{{pos}_{i}}} \\
    & \frac{\partial \log(\mathcal{G}_{neg}(\mathbf{x}))}{\partial x_i} = -\frac{x_{i} - \mu_{{neg}_{i}}}{\sigma^2_{{neg}_{i}}} \notag
\end{align}

Eventually, the gradients in one batch are computed as follows:
\begin{equation}
    \frac{\partial \mathcal{S}}{\partial x_i} = \sum_{j=1}^M \frac{\partial \mathcal{S}}{\partial x^j_i}
\end{equation}
where $M$ is the batch size of current frame and $x^j_i$ is the $i$th element of the $j$th image, $\mathbf{x}^j$, in one batch. 

To avoid updating the parameters aggressively, during online tracking the parameters are only updated if the tracking system is confident about the new appearance of the target. The confidence value is proportional to the tracking score, $\mathcal{S}(\mathbf{x})$. This strategy makes the tracker capable of recovering the target when an occlusion happens.

\begin{figure*}[!tph]
  \begin{minipage}[b]{0.49\textwidth}
  \centering
  \begin{subfigure}[b]{0.49\textwidth}
  {\includegraphics[width=\textwidth]{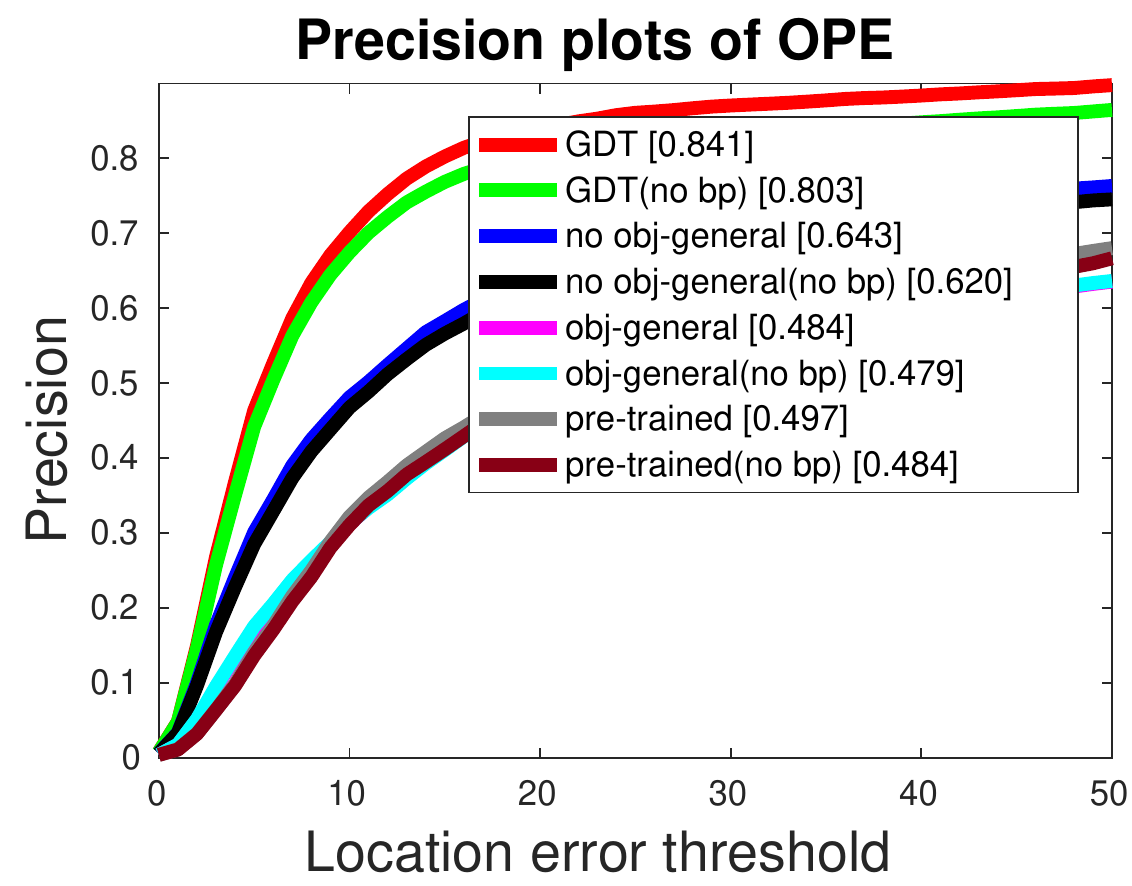}}
  \label{fig:Precision_AblationStudy}
  \end{subfigure}
  \centering  
  \begin{subfigure}[b]{0.49\textwidth}
  {\includegraphics[width=\textwidth]{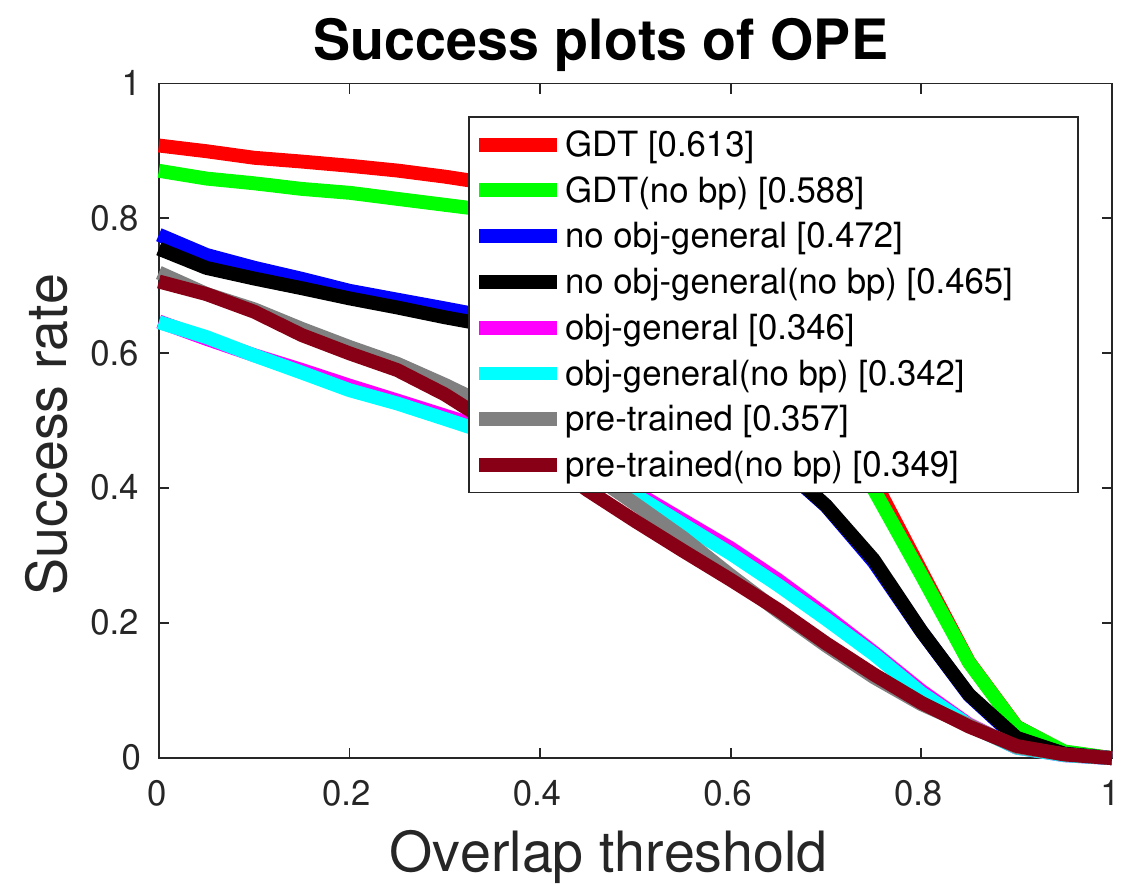}}
  \label{fig:Success_AblationStudy}
  \end{subfigure}
  \caption{Ablation Study}
  \label{fig:AblationStudy}
  \end{minipage}
  \begin{minipage}[b]{0.49\textwidth}
  \centering  
  \begin{subfigure}[b]{0.49\textwidth}
  {\includegraphics[width=\textwidth]{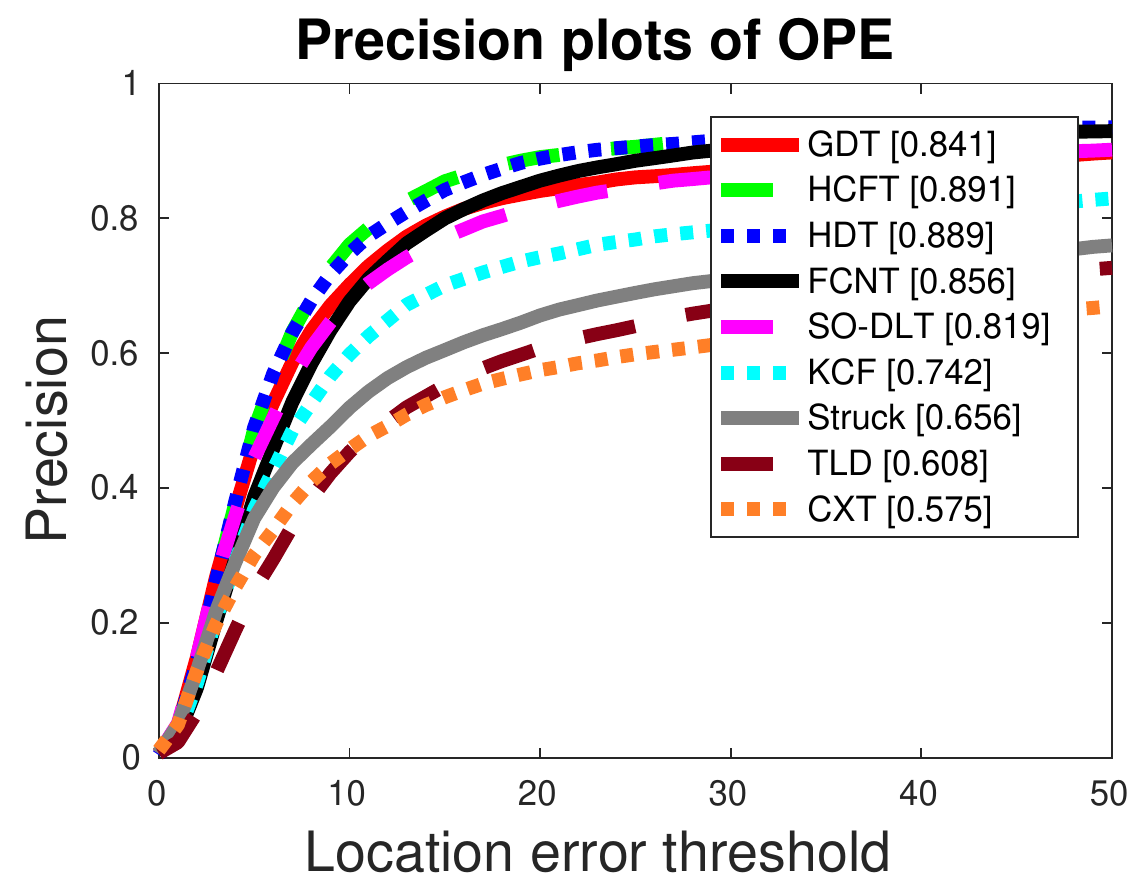}}
  \label{fig:PrecisionPlot}
  \end{subfigure} 
  \centering  
  \begin{subfigure}[b]{0.49\textwidth}
  {\includegraphics[width=\textwidth]{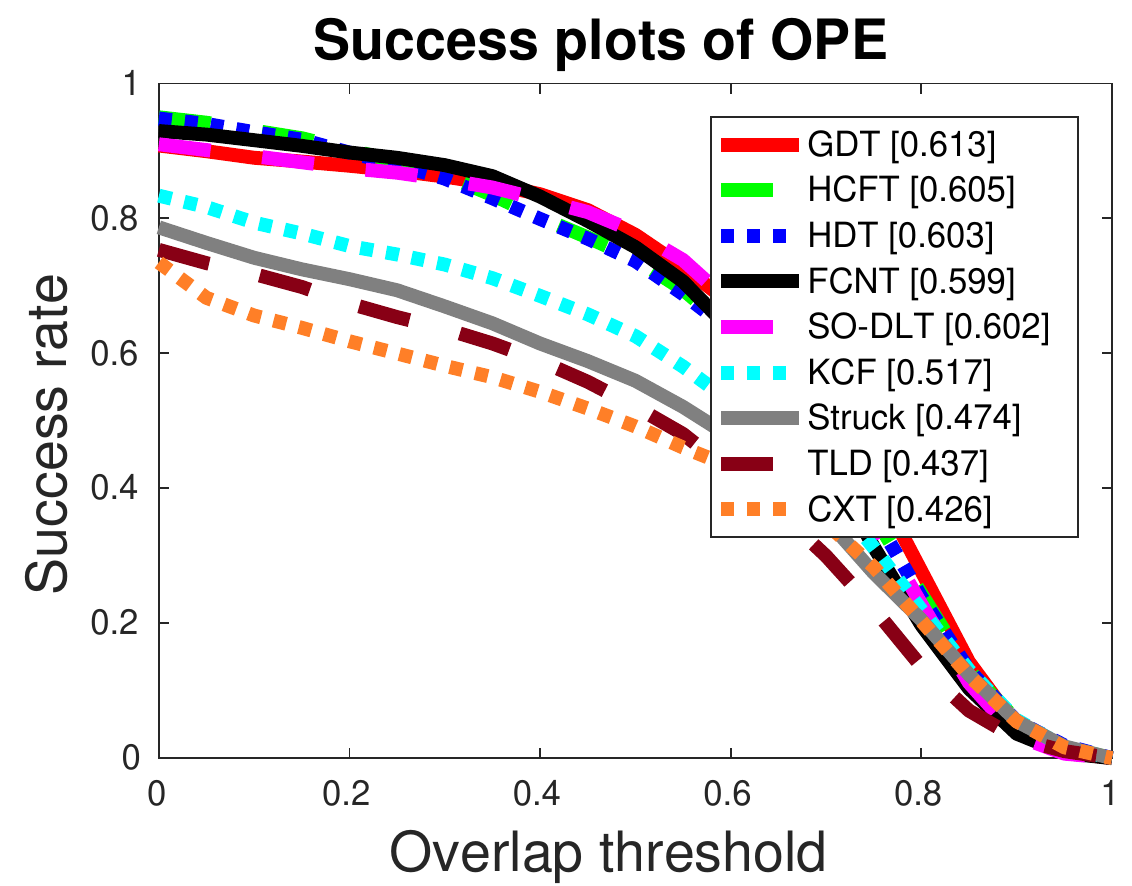}}
  \label{fig:SuccessPlot}
  \end{subfigure}
  \caption{Quantitative Results}
  \label{fig:SuccessPrecisionPlots}
  \end{minipage}
\end{figure*}

In summary, the online tracking algorithm is a two-stage iterative process in which the network parameters are updated to maintain the distributions of negative and positive examples. The algorithm starts with an initial set of means and variances estimated from the bounding boxes in the first frame. Then when a new frame comes, the following steps are performed:
\begin{enumerate}
\item Forward - Stage 1. In the forward procedure of the CNN, given a fine-tuned neural network, we find the new location which has the highest score and re-estimate Gaussian parameters. Estimation of the Gaussian parameters is a deterministic procedure which uses maximum likelihood estimation.
\item Backward - Stage 2. In the backward procedure of the CNN, the Gaussian parameters are fixed and the gradients of the tracking score, $\mathcal{S}$, with respect to $\mathbf{x}$ are computed in order to propagate tracking error to the network and update the CNN parameters. While backpropogation, only fully connected layers are updated.
\end{enumerate}

\section{Experiments}
In order to evaluate the performance of our deep learning based tracker, we have carried out extensive experiments using the CVPR13 ``Visual Tracker Benchmark'' dataset~\cite{WuLimYang13}. It contains 50 challenging video sequences from complex scenes and covers a diverse set of visual attributes including illumination variation (IV), occlusion (OCC), scale variation (SV), deformation (DEF), motion blur (MB), fast target motion (FM), in-plane and out of plane rotations (IPR and OPR), out-of-view (OV), background clutter (BC), and low resolution videos (LR).  These are the typical challenges for visual tracking systems. We follow the ``Visual Tracker Benchmark'' protocol introduced in~\cite{WuLimYang13} in order to compare the tracking accuracy to the state-of-the-art approaches. 

Following the evaluation protocol in~\cite{WuLimYang13}, the experimental results are illustrated in terms of both {\em precision} and {\em success}. The precision plot shows the percentage of the frames in which the target is tracked. The center location error between the tracked target and ground truth is measured at different threshold values. The representative precision score is computed at the threshold value equal to $20$ pixels. Similarly, the success plot shows the percentage of frames in which the target is successfully tracked. This is done by measuring the overlap ratio of a prediction bounding box with the ground truth one as the intersection over union, and applying different threshold values between $0$ and $1$. We run the One-Pass Evaluation (OPE) on the benchmark and use the online available code provided by~\cite{WuLimYang13} to generate the evaluation plots.

In our experiments, Opencv\footnote{\url{http://opencv.org}} and Caffe\footnote{\url{http://caffe.berkeleyvision.org}} libraries are used for the CNN-based tracking system. The CNN is fine-tuned for $100k$ iterations for objectness and the maximum number of iterations for the specific target fine-tuning in the first frame is set to be equal to $500$. During online tracking, the CNN is backpropogated $1$ iteration per frame. The aspect ratio is fixed as the same as the initialization given in the first frame of each sequence. The number of scales in the image pyramid is set to be equal to $3$ and the scale step is equal to $0.02$\footnote{If the scale step is not enough to change the size of the bounding box by at least one pixel, it is increased gradually to achieve that amount of change.}. The learning rate for Gaussian parameters is set to $0.95$. During online tracking, the parameters of fully connected layers are only updated when the tracking score is not negative and similarities between current frame and last updated frame is above a similarity threshold. The current prototype of the proposed algorithm runs at approximately 1 fps on a PC with an Intel i7-4790 CPU and a Nvidia Titan X GPU.

\subsection{Ablation Studies}
In order to observe the effectiveness of each weight-tuning step in our algorithm we have conducted multiple experiments with three pairs of baselines. The first pair of baseline, which we refer to it as the "pre-trained" is to take the pre-trained model~\cite{NIPS2012_4824} as the feature extractor (without fine-tuning for objectness and target appearance) and use the same tracker as GDT to track every target in each sequence. By "no bp" we mean that during tracking process only Gaussian parameters are updated and CNNs are not fine-tuned. The second pair of baselines, which we call them the "obj-general", is to take the CNN model we trained for objectness as the feature extractor. To show the importance of fine-tuning for objectness, we add third pair of baselines, which we refer to as the "no obj-general". For this baseline, we remove the objectness step and CNNs are fine-tuned directly from the pre-trained model. All results listed in this section adopt same tracker, the only difference is the CNN models that are used. We summarize comparisons with all baselines in \figurename{~\ref{fig:AblationStudy}} and more details are listed in \tablename{~\ref{tab:AblationDetails}}. From \figurename{~\ref{fig:AblationStudy}}, it is clear that each step of our algorithm boosts the tracking results.

\begin{table*}
\caption{Ablation Study.}
\label{tab:AblationDetails}
\begin{center}
\small{
\begin{tabular}{|c|c|c|c|}
\hline
Method & Precision Score & Success Score & feature dimension \\ \hline
GDT & 0.841 & 0.613 & 500\\ \hline
GDT(no bp) & 0.803 & 0.588 & 500\\ \hline
no obj-general & 0.643 &  0.472 & 500\\ \hline
no obj-general(no bp) & 0.620 &  0.465 & 500\\ \hline
obj-general & 0.497 &  0.357 & 500\\ \hline
obj-general(no bp) & 0.484 &  0.349 & 500\\ \hline
pre-trained & 0.484 &  0.346 & 4096\\ \hline
pre-trained(no bp) & 0.479 &  0.342 & 4096\\ \hline
\end{tabular}
}
\end{center}
\end{table*}

Firstly, as we can see from ablation studies, removing fine-tuning for objectness results in a large drop of tracking results. Since for the tracking problem the amount of available training data is very limited (one training example from the first frame in each sequence), fine-tuning with auxiliary data is very important. However, the CNN trained for objectness itself does not bring any improvement on tracking since both obj-general and pre-trained models do not contain any feature learning for certain tracking targets. In other words, objectness greatly contributes to the later fine-tuning steps. Secondly, obj-specific fine-tuning largely boost the tracking results. The benefit of this step is obvious, the CNN is trained for a certain tracking target and the learnt features are more discriminative. The ablation study also suggest that online fine-tuning does have a positive impact on tracking results which means learning object features adaptively during tracking is an important step in our deep learning tracking system.

\subsection{Quantitative Results}
\label{sec:results_part2}

Our tracking results are quantitatively compared with the eight state state-of-the-art tracking algorithms with the same initial location of the target. These algorithms are tracking-by-detection (TLD)~\cite{Kalal2010PNclassifiersTracking}, context tracker (CXT)~\cite{dinh2011context}, Struck~\cite{HareSTRUCK2011}, kernelized correlation filters (KCF)~\cite{Henriques2015KCF}, structured output deep learning tracker (SO-DLT)~\cite{wang2015transferring}, fully convolutional
network based tracker (FCNT)~\cite{wang2015visual}, hierarchical convolutional features for visual tracking (HCFT)~\cite{ma2015hierarchical}, and hedged deep tracking (HDT)~\cite{Qi2016HedgedDT}. The first four algorithms are among the best trackers in the literature which use hand-crafted features, and the last four are among best approaches for CNN-based tracking. 

\figurename{~\ref{fig:SuccessPrecisionPlots}} shows the success and precision plots for the whole $50$ videos in the dataset. Overall, the proposed tracking algorithm performs favorably against the other state-of-the-art algorithms on all tested sequences. It outperforms all of the state-of-the-art approaches given success plot and produces favourable results compared to other deep learning-based trackers given precision plot, specifically for low location error threshold values. \tablename{~\ref{tab:TrackingScores}} summarizes the tracking scores for state-of-the-art trackers, the best method is highlighted in red color.

\begin{table}
\caption{Average scores for different trackers.}
\label{tab:TrackingScores}
\begin{center}
\small{
\begin{tabular}{|c|c|c|}
\hline
Method & Precision Score & Success Score \\ \hline
GDT(Ours) & 0.841 & \textbf{\textcolor{red}{0.613}}\\ \hline
FCNT~\cite{wang2015visual} & 0.856 & 0.599 \\ \hline
HCFT~\cite{ma2015hierarchical} & \textbf{\textcolor{red}{0.891}} & 0.605 \\ \hline
HDT~\cite{Qi2016HedgedDT} & 0.889 & 0.603 \\ \hline
SO-DLT~\cite{wang2015transferring} & 0.819 & 0.602\\ \hline
KCF~\cite{Henriques2015KCF} & 0.742 &  0.517\\ \hline
Struck~\cite{HareSTRUCK2011} & 0.656 &  0.474\\ \hline
TLD~\cite{Kalal2010PNclassifiersTracking} & 0.608 &  0.437\\ \hline
CXT~\cite{dinh2011context} & 0.575 &  0.426\\ \hline
\end{tabular}
}
\end{center}
\end{table}

\begin{figure}[!tph] 
  \centering  
  \begin{subfigure}[b]{0.45\textwidth}
  {\includegraphics[width=\textwidth]{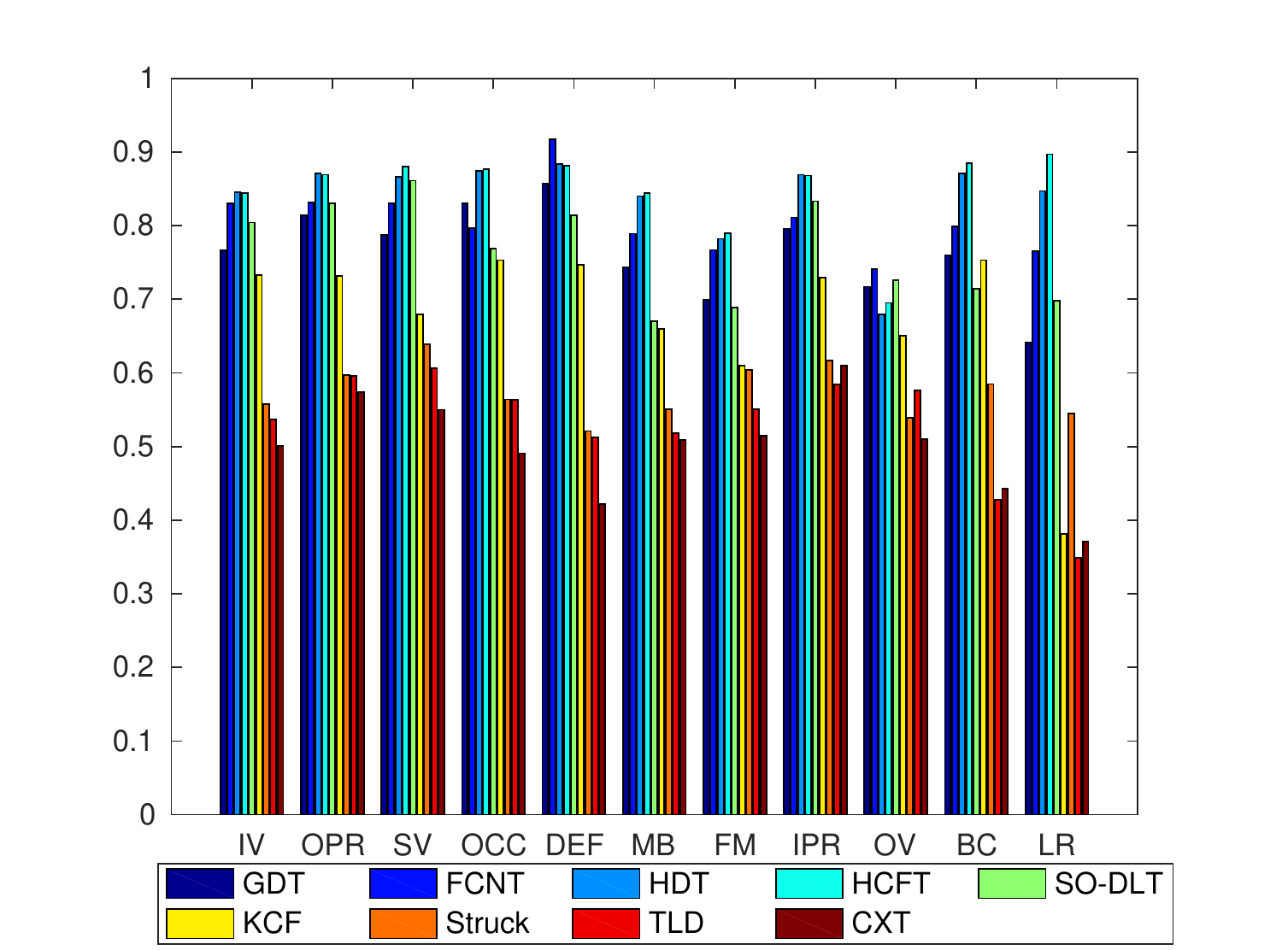} } \vspace{-10pt}
  \caption{Precision score for different attributes}
  \label{fig:PrecisionAttribute}
  \end{subfigure} 
  \hfill
  \centering  
  \begin{subfigure}[b]{0.45\textwidth}
  {\includegraphics[width=\textwidth]{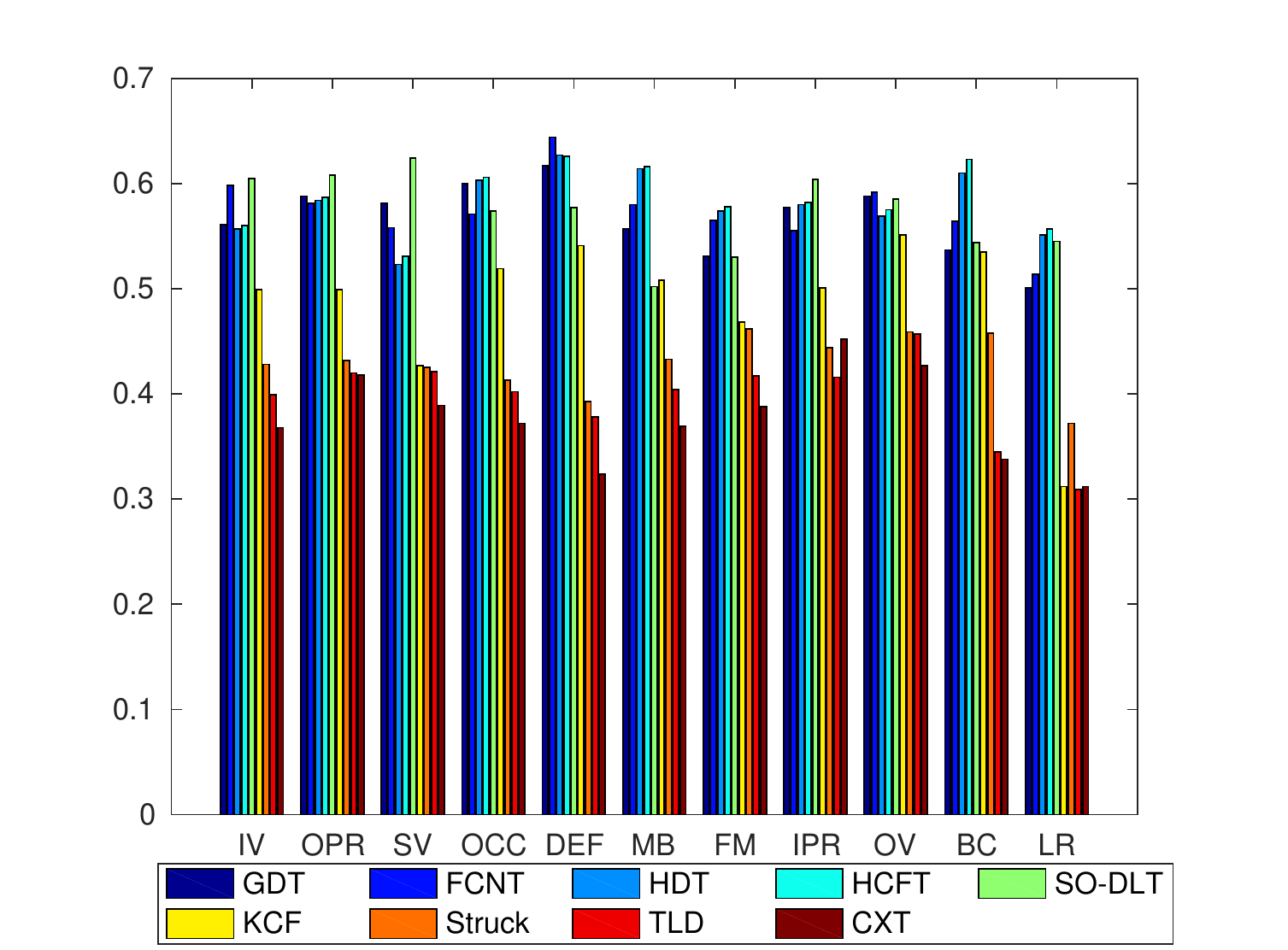} } \vspace{-10pt}
  \caption{Success score for different attributes}
  \label{fig:SuccessAttribute}
  \end{subfigure}   
  \caption{Precision and success scores on different attributes.}
  \label{fig:AttributePlots}
\end{figure}

\begin{figure*}
	\centering	
	\begin{subfigure}[b]{0.24\textwidth}
	{\includegraphics[width=\textwidth]{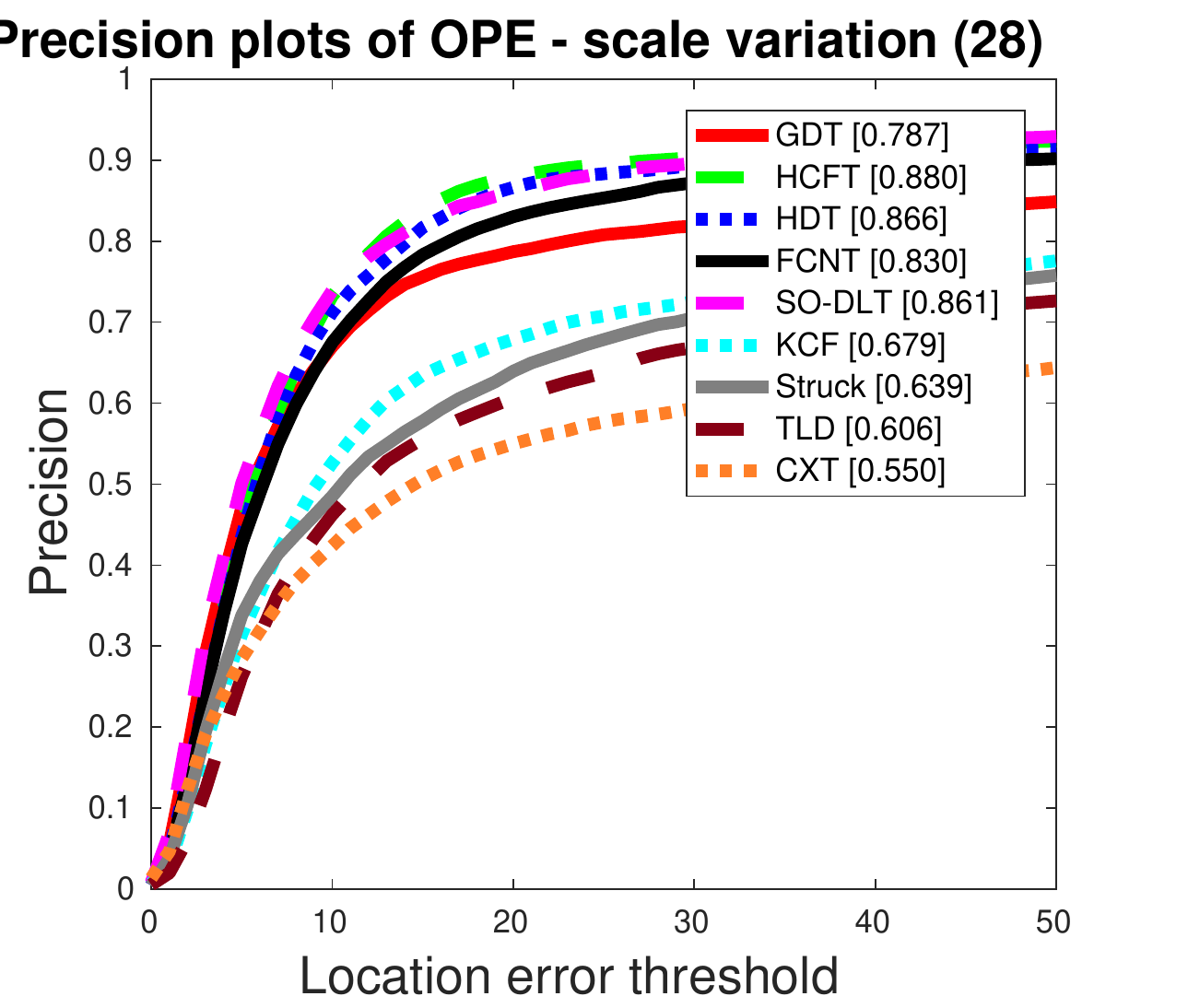}}
	\end{subfigure} 
	\centering	
	\begin{subfigure}[b]{0.24\textwidth}
	{\includegraphics[width=\textwidth]{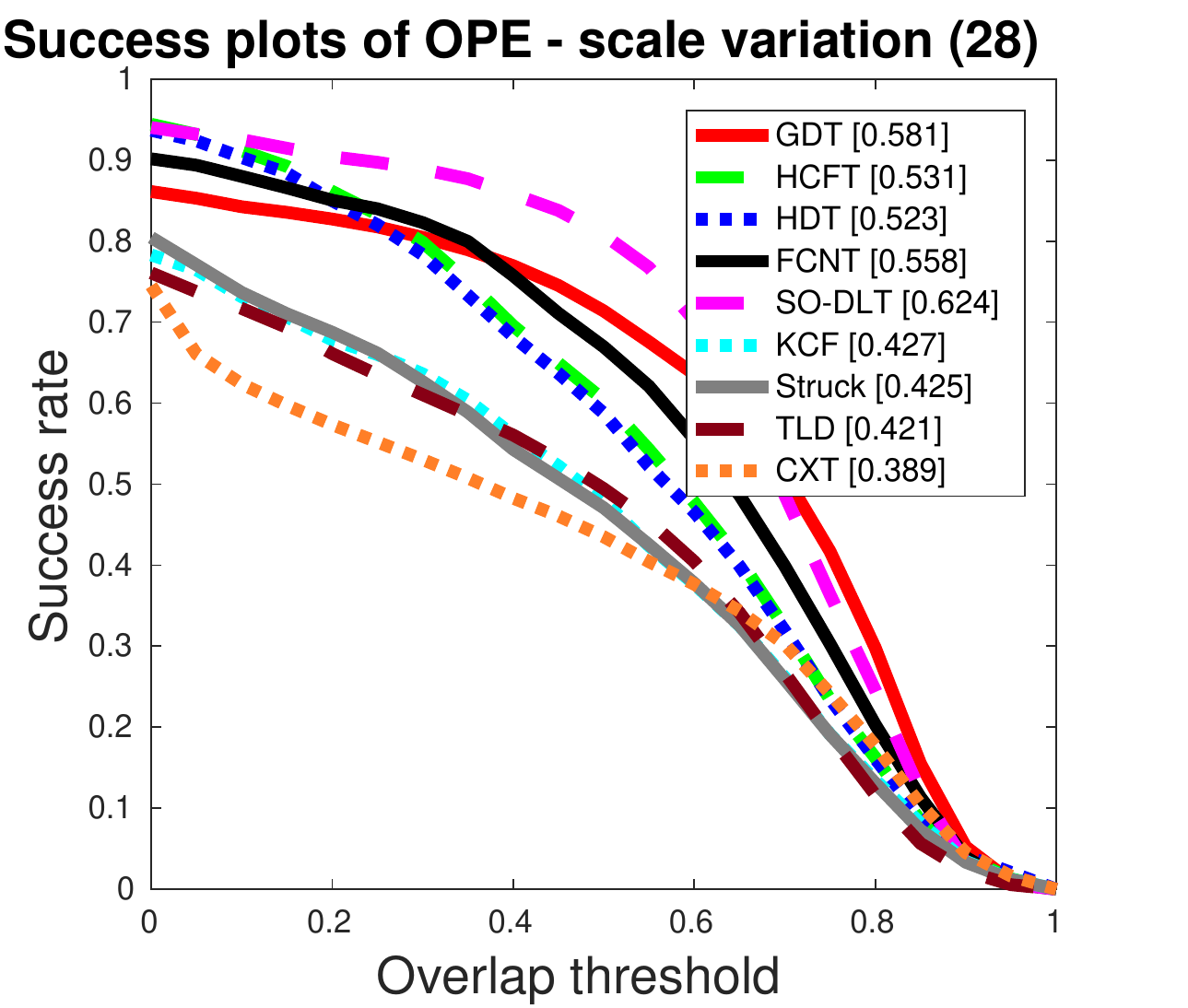}}
	\end{subfigure} 
	\centering	
	\begin{subfigure}[b]{0.24\textwidth}
	{\includegraphics[width=\textwidth]{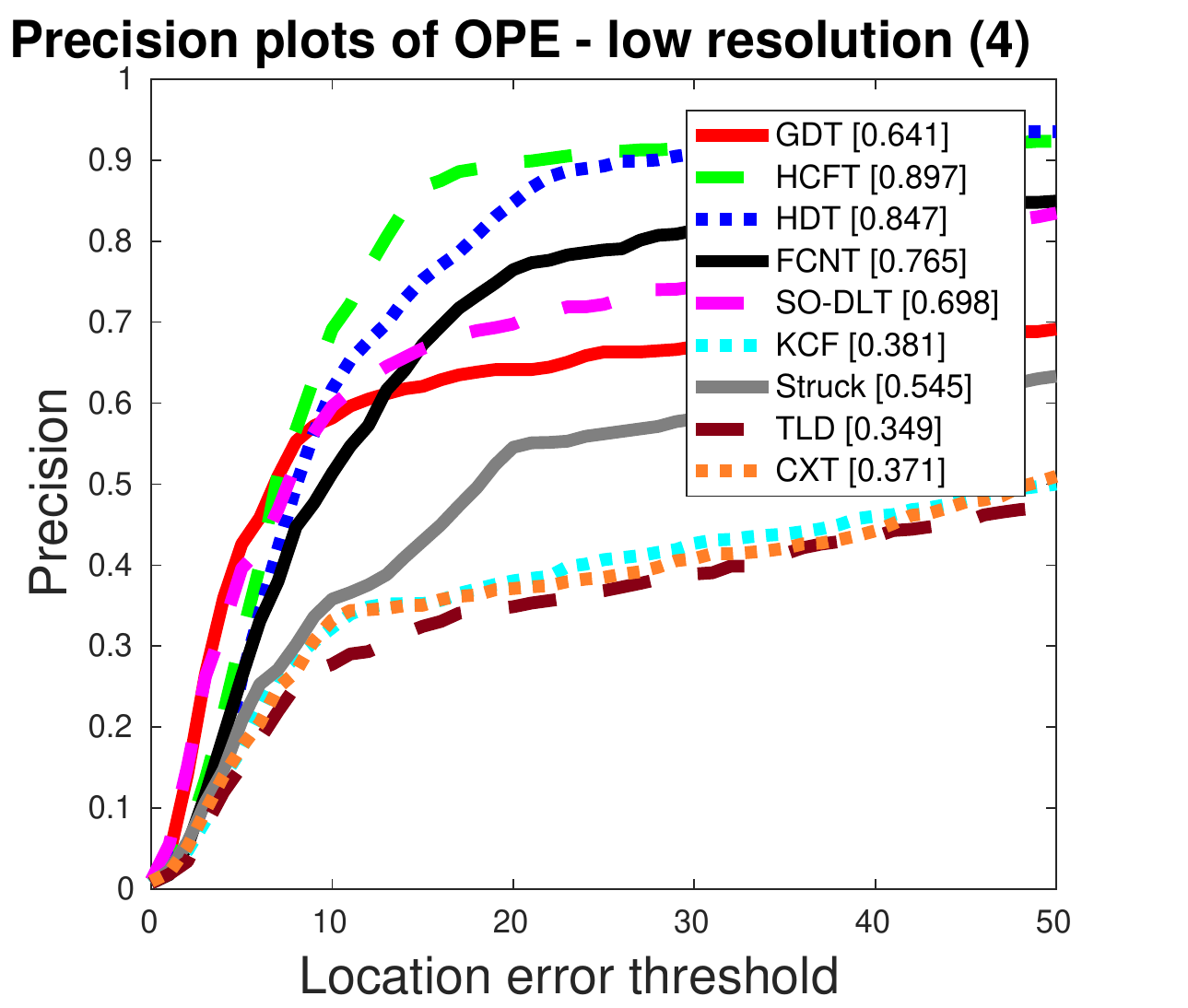}} 
	\end{subfigure}
	\centering	
	\begin{subfigure}[b]{0.24\textwidth}
	{\includegraphics[width=\textwidth]{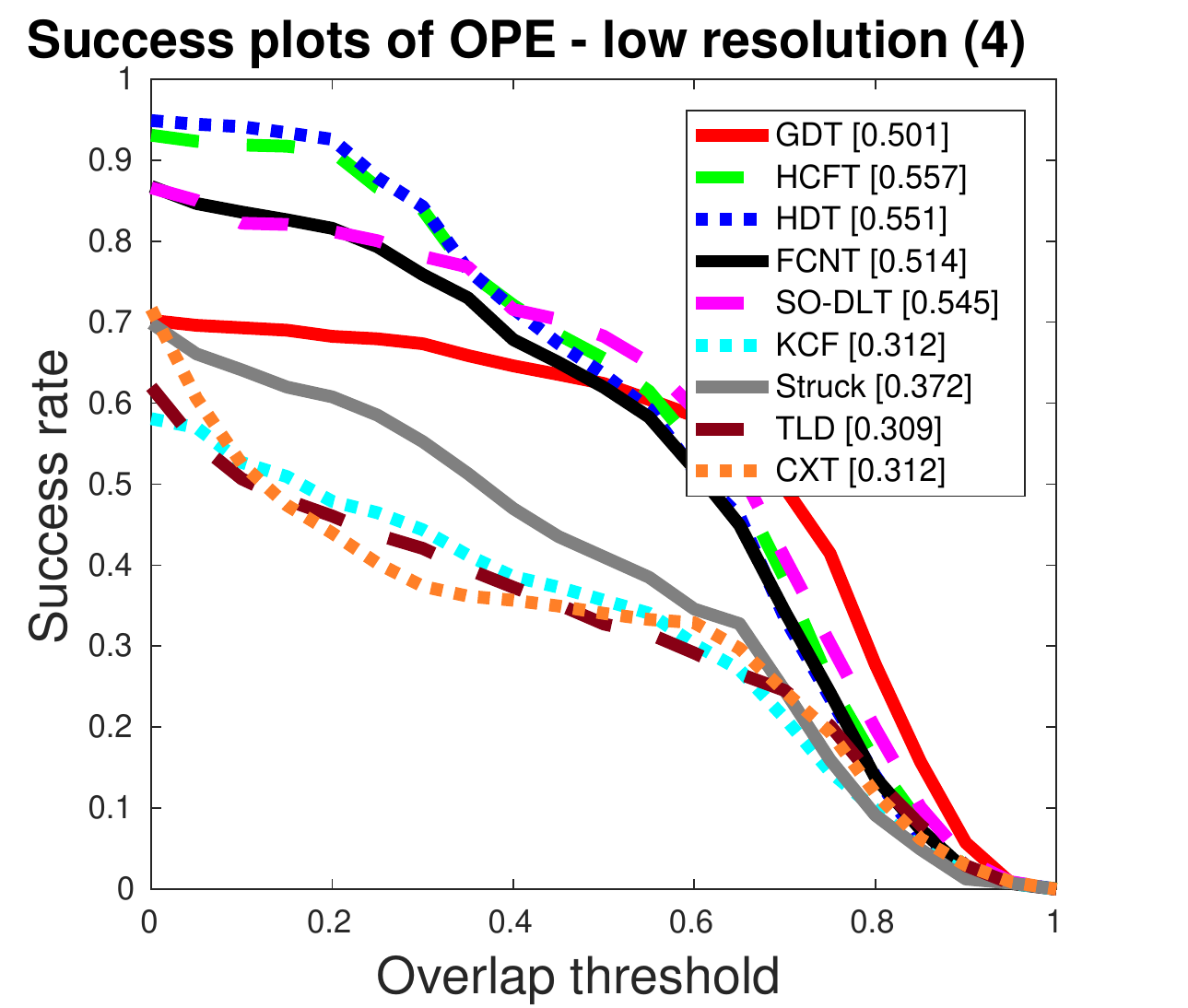}}
	\end{subfigure} \\
	\caption{Precision and success plots for individual attributes: scale variation (SV) and low resolution (LR). }
	\label{fig:AttributeIndividualPlots}
\end{figure*}

\begin{figure*}[!tph] 
\centering
\begin{subfigure}[b]{0.18\textwidth}
\includegraphics[width=\textwidth]{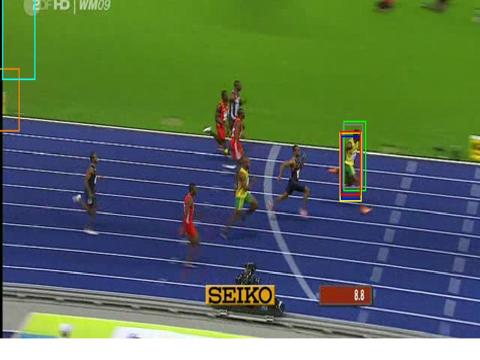}
\caption*{Bolt}
\end{subfigure}
\begin{subfigure}[b]{0.18\textwidth}
\includegraphics[width=\textwidth]{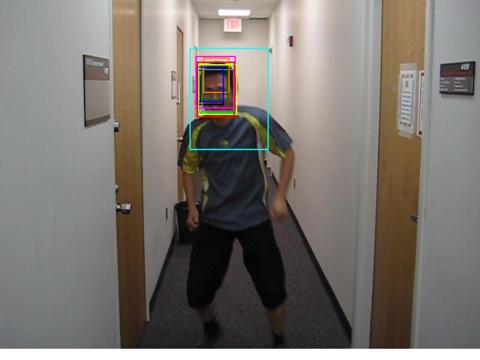} 
\caption*{Boy}
\end{subfigure}
\begin{subfigure}[b]{0.18\textwidth}
\includegraphics[width=\textwidth]{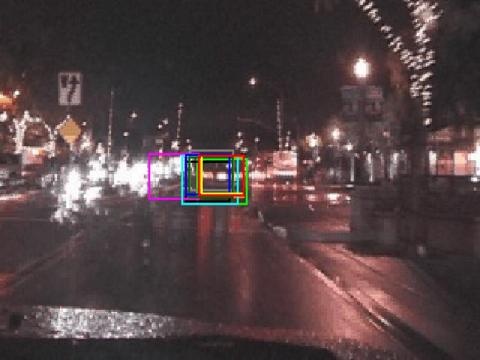}
\caption*{carDark}
\end{subfigure}
\begin{subfigure}[b]{0.18\textwidth}
\includegraphics[width=\textwidth]{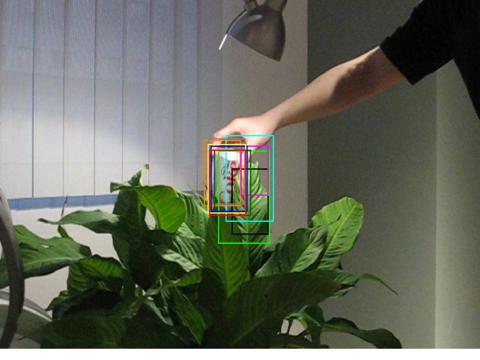} 
\caption*{Coke}
\end{subfigure}
\begin{subfigure}[b]{0.18\textwidth}
\includegraphics[width=\textwidth]{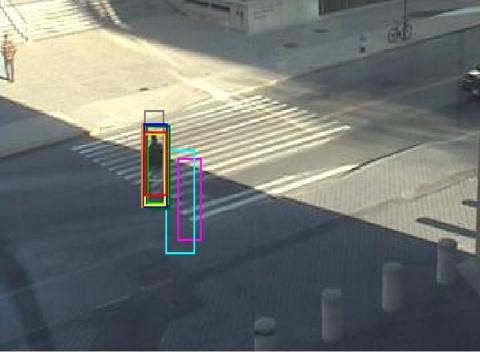}
\caption*{Crossing}
\end{subfigure} \\

\begin{subfigure}[b]{0.18\textwidth}
\includegraphics[width=\textwidth]{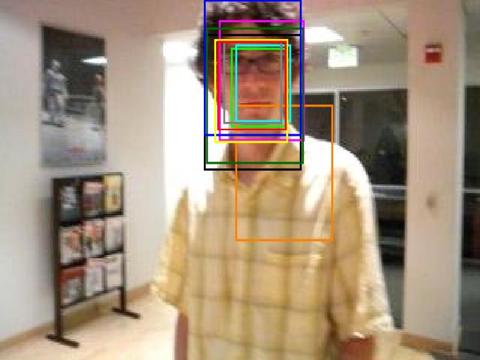}
\caption*{David}
\end{subfigure}
\begin{subfigure}[b]{0.18\textwidth}
\includegraphics[width=\textwidth]{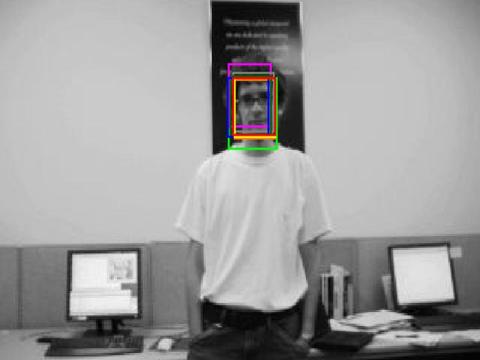}
\caption*{David2}
\end{subfigure} 
\begin{subfigure}[b]{0.18\textwidth}
\includegraphics[width=\textwidth]{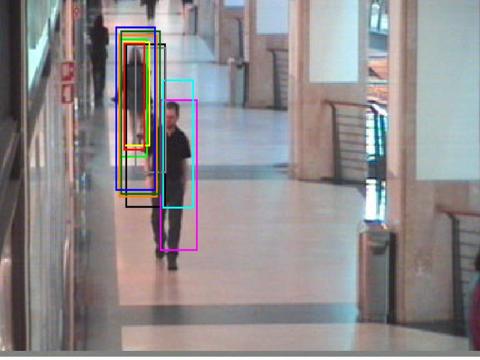}
\caption*{Walking2}
\end{subfigure}
\begin{subfigure}[b]{0.18\textwidth}
\includegraphics[width=\textwidth]{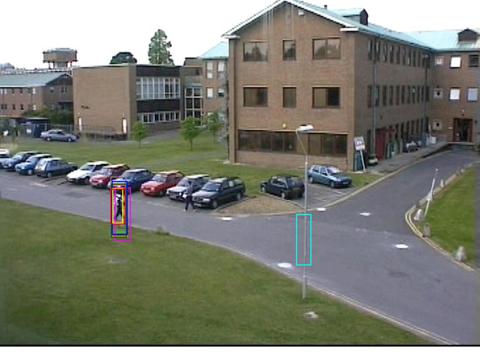}
\caption*{Walking}
\end{subfigure}
\begin{subfigure}[b]{0.18\textwidth}
\includegraphics[width=\textwidth]{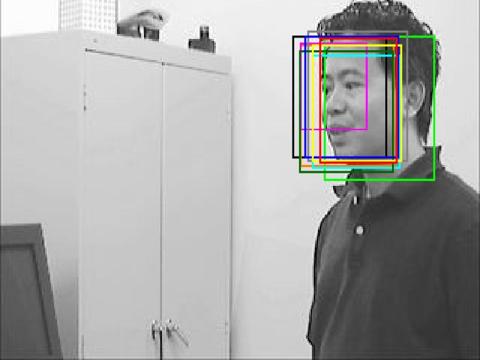}
\caption*{Dudek}
\end{subfigure} \\

\begin{subfigure}[b]{0.18\textwidth}
\includegraphics[width=\textwidth]{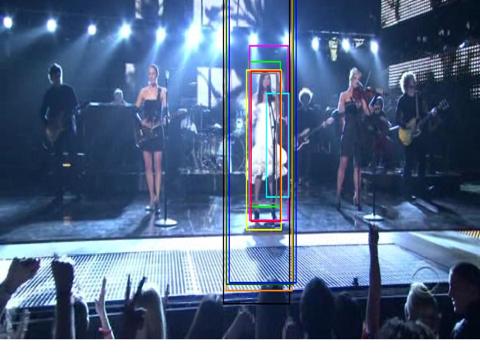}
\caption*{Singer2}
\end{subfigure}
\begin{subfigure}[b]{0.18\textwidth}
\includegraphics[width=\textwidth]{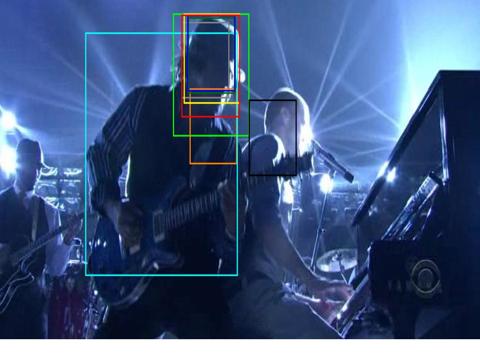}
\caption*{Shaking}
\end{subfigure}
\begin{subfigure}[b]{0.18\textwidth}
\includegraphics[width=\textwidth]{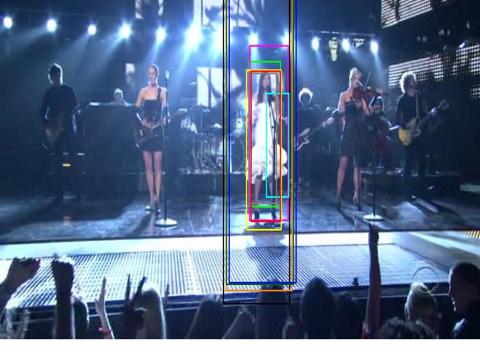}
\caption*{Singer1}
\end{subfigure}
\begin{subfigure}[b]{0.18\textwidth}
\includegraphics[width=\textwidth]{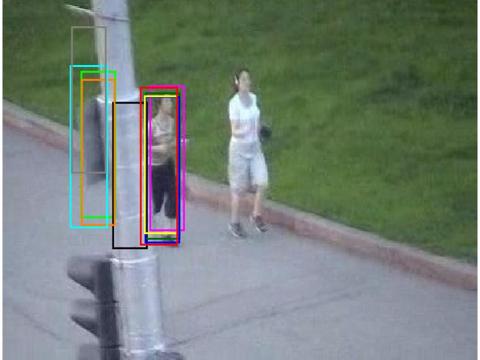}
\caption*{Jogging1}
\end{subfigure}
\begin{subfigure}[b]{0.18\textwidth}
\includegraphics[width=\textwidth]{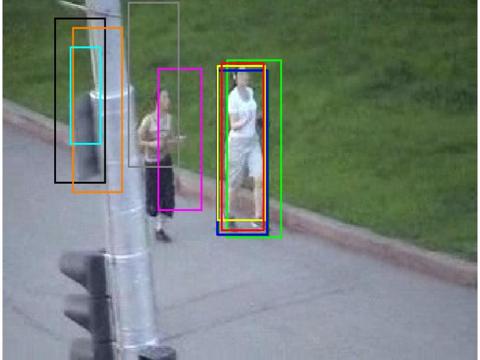}
\caption*{Jogging2}
\end{subfigure} \\
\caption{Visualizations of all tracking algorithms on challenging sequences. Ground Truth: red, GDT(ours): yellow, FCNT: gray, HDT: dark green, HCFT: blue, SO-DLT: green, KCF: black, Struck: orange, TLD: magenta, CXT: cyan.}
\end{figure*}

In order to have a more detailed comparison, the success rate and precision scores are reported for different tracking attributes in \figurename{~\ref{fig:AttributePlots}}. It is clear that the proposed tracker outperforms all of the non-deep learning based tracking systems in every single attribute. The state-of-the-art deep learning based trackers, FCNT and SO-DLT, show similar performance in some attributes and the other two deep learning based trackers, HDT and HCFT, show better performance in most attributes. It is somewhat expected because both HDT and HCFT trackers take advantage of multiple convolutional layers compared to ours. However, despite their high accuracy in terms of precision, their success score is lower than our algorithm. In other words, considering the center location of the target, HDT and HCFT can track the target better than the proposed GDT algorithm while GDT demonstrates better capabilities in localizing the target. More specifically, the GDT method can localize the target with a higher accuracy in the \emph{out-of-view} (OV) test scenario where the target is invisible for a long period of time. 
The reason is that our approach does not estimate the object location but instead treat object locations and scales as a whole, while they infer object locations from each convolutional layers. This is more obvious in the \emph{scale variation} (SV) and \emph{low resolution} (LR) attributes where the success scores drop dramatically compared with the precision scores (see \figurename{~\ref{fig:AttributeIndividualPlots}}). Given the success and precision plots for the LR attribute, it is obvious that our tracker has a higher accuracy for small amounts of location error and high overlap ratios. On the other hand, the discriminatory power of the estimated distributions of the learnt features seems to be more effective in learning the appearance variations of a target and hence, our tracker shows good performance in \emph{occlusions} (OCC) and \emph{deformation} (DEF) attributes as shown in \figurename{~\ref{fig:AttributePlots}}.

The good performance of our proposed algorithm results from two main reasons. The first reason is that the CNN model pre-trained on an auxiliary dataset provides more effective features than conventional hand-crafted features and produce results comparable to approaches using features extracted from multiple convolutional layers. After the second step of training using bounding boxes from the first frame of each sequence, the features are more target-specific. The effectiveness of the learnt strong appearance feature can be justified by the results on the sequences with appearance changes, e.g. the {\em deformation} attribute.
The second reason is that the way that we update our model makes it more robust to appearance changes and occlusions. The Gaussian update strategy lets our model have memory for previous tracking results, while obviating the need for two-stream approaches for storing previous features. The effectiveness of the updating strategy can be seen from the results on the sequences with \emph{out of view} and \emph{occlusion}. Overall, stronger appearance features learnt for each target combined with an update strategy makes the proposed tracker capable of accurately tracking and localizing the target.


\section{Conclusion}
We proposed a novel tracking algorithm in this paper. The CNN for tracking is trained in a simple but very effective way and the CNN provides good features for object tracking. First stage fine-tuning using auxiliary data largely alleviates the problem of a lack of labelled training instances. A second stage of fine-tuning, though used only with a few hundred instances and trained for tens of iterations, greatly boosts the performance of the tracker. On top of CNN features, a classifier is learnt by adding another layer on top of the fc7 layer. The experimental results show that our deep, appearance model learning tracker produces results comparable to state-of-the-art approaches and can generate accurate tracking results.

\ifCLASSOPTIONcaptionsoff
  \newpage
\fi

\bibliographystyle{IEEEtran}
\bibliography{TrackingRefs}

\end{document}